\pdfoutput=1

\documentclass[11pt]{article}
\usepackage{authblk}
\usepackage{acl}
\usepackage{times}
\usepackage{latexsym}

\usepackage[T1]{fontenc}

\usepackage[utf8]{inputenc}

\usepackage{microtype}

\usepackage{graphicx}
\usepackage{tabularx}
\newcolumntype{Y}{>{\centering\arraybackslash}X}

\usepackage{hyperref}
\usepackage{xstring}
\usepackage{color}
\usepackage{multicol}
\usepackage{multirow}
\usepackage{hhline}
\usepackage{mathtools}
\usepackage[nomessages]{fp}
\usepackage{makecell}

\usepackage{amsmath}

\DeclareMathOperator{\arccosh}{arcCosh}

\title{Sinhala Sentence Embedding: A Two-Tiered Structure for Low-Resource Languages}

\author[*,1]{\textbf{Gihan Weeraprameshwara}}
\author[*]{\textbf{Vihanga Jayawickrama}}
\author[*]{\textbf{Nisansa de Silva}}
\author[**]{\authorcr \textbf{Yudhanjaya Wijeratne}}
\affil[*]{Department of Computer Science \& Engineering, University of Moratuwa, Sri Lanka}
\affil[**]{LIRNEasia, Sri Lanka}
\affil{{\tt 
gihanravindu.17@cse.mrt.ac.lk}}

\begin{document}
\maketitle
\begin{abstract}
In the process of numerically modeling natural languages, developing language embeddings is a vital step. However, it is challenging to develop functional embeddings for resource-poor languages such as Sinhala, for which sufficiently large corpora, effective language parsers, and any other required resources are difficult to find. In such conditions, the exploitation of existing models to come up with an efficacious embedding methodology to numerically represent text could be quite fruitful. This paper explores the effectivity of several one-tiered and two-tiered embedding architectures in representing Sinhala text in the sentiment analysis domain. With our findings, the two-tiered embedding architecture where the lower-tier consists of a word embedding and the upper-tier consists of a sentence embedding has been proven to perform better than one-tier word embeddings, by achieving a maximum F1 score of 88.04\% in contrast to the 83.76\% achieved by word embedding models. Furthermore, embeddings in the hyperbolic space are also developed and compared with Euclidean embeddings in terms of performance. A sentiment data set consisting of Facebook posts and associated reactions have been used for this research. To effectively compare the performance of different embedding systems, the same deep neural network structure has been trained on sentiment data with each of the embedding systems used to encode the text associated.
\end{abstract}

\section{Introduction}
\label{sec:intro}

An effective numerical representation of the textual content is crucial for natural language processing models, in order to understand the underlying relational patterns among words and discover patterns in natural languages. For resource-rich languages like English, numerous pre-trained models as well as the required materials to develop an embedding system are readily available. On the contrary, for resource-poor languages such as Sinhala, neither of those options could be easily found \cite{de2019survey}. Even the data sets that are available for training often fail to meet adequate standards \cite{caswell2021quality}. Thus, discovering a convenient methodology to develop embeddings for text would be a great step forward in the NLP domain for the Sinhala language. 

\begin{figure}[!htb]
\centering
\includegraphics[width=0.35\textwidth]{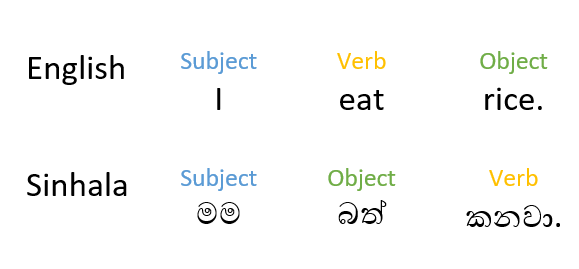}
    \caption{SVO grammar structure of English and SOV grammar structure of Sinhala}
    \label{fig:grammar}
\end{figure}

Sinhala, also known as Sinhalese, is an Indo-Aryan language that is used within Sri Lanka \cite{kanduboda2011role}. The primary user base of this language is the Sinhalese ethnic group of the country. In total, 17 million people use Sinhala as their first language while 2 million people use it as a second language \cite{de2019survey}. Furthermore, Sinhala is structurally different from English, which uses a subject-verb-object structure as opposed to the subject-object-verb structure used by Sinhala as shown in the figure~\ref{fig:grammar} thus most of the pre-trained embedding models for English may not be effective with Sinhala. 

This study therefore is focused on discovering an effective embedding system for Sinhala text that provides reasonable results when used in training deep learning models. Sentiment analysis with Facebook data is utilized as the use case for the study.

Upon considering common forms of vector presentations of textual content, bag of words, word embedding, and sentence embedding are three of the leading methodologies in the present. Word embeddings have been observed to surpass the performance of bag of words for large enough data sets \cite{rudkowsky2018more} because bag of words often met with various problems such as disregarding the grammatical structure of the text, large vocabulary dimension and sparse representation \cite{le2014distributed,el2016enhancement}. In order to tackle the above challenges, word embeddings can be used. Since word embeddings capture the similarities among ingrained sentiments in words and represent them in the vector space, word embeddings tend to increase the accuracy of classification models \cite{goldberg2016primer}.

However, one of the major weaknesses of word embedding models is that they fail to capture syntax and polysemy; i.e. the presence of multiple possible meanings for a certain word or a phrase \cite{mu2016geometry}. In order to overcome these obstacles and also to achieve fine granularity in the embedding, sentence embeddings are used. The idea is to test common Euclidean space word embedding techniques such as fastText \cite{bojanowski2017enriching,joulin2016bag}, Word2vec \cite{mikolov2013efficient}, and GloVe \cite{pennington2014glove} with sentence embedding techniques. The pooling methods (i.e. max pooling, min pooling and avg pooling) will be considered as the baseline methods for the test. More advanced models such as sequence to sequence model (i.e. seq2seq model) \cite{sutskever2014sequence} and the modified version of the sequence to sequence model introduced by the work of~\newcite{cho2014learning} with GRU \cite{chung2014empirical} and LSTM \cite{hochreiter1997long} recurrent neural network units will be tested against the pooling means. Furthermore, the addition of attention mechanism \cite{vaswani2017attention} into the sequence to sequence model will also be tested. 

Most models created using word and sentence embeddings are based on the Euclidean space. Though this vector space is commonly used, it poses significant limitations when representing complex structures \cite{nickel2017poincare}. Using the hyperbolic space provides a plausible solution for such instances. The hyperbolic space is a negatively-curved, non-Euclidean space. It is advantageous for embedding trees as the circumference of a circle grows exponentially with the radius. The usage of hyperbolic embedding is still a novel research area as it was only introduced recently, through the work of~\newcite{nickel2017poincare,chamberlain2017neural,sala2018representation}. The work of~\newcite{lu2019learning,lu2020exploiting} highlight the importance of using the hyperbolic space to improve the quality of embeddings in a practical context within the medical domain. However, research done on the applicability of hyperbolic embeddings in different arenas is highly limited. Thus, the full potential of the hyperbolic space is yet to be fully uncovered. 

Through this paper, we are testing the effectiveness of a set of two-tiered word representation models that include various word embeddings as the lower tier and sentence embeddings as the upper tier will be compared.

\section{Related Work}
\label{sec:related}
The sequence to sequence model introduced by the work of~\newcite{sutskever2014sequence} is vital in this research as it is one of the core models in developing sentence embedding. Though originally developed for translation purposes the model has gone under multiple modifications depending on the context such as description generation for images \cite{karpathy2015deep}, phrase representation \cite{cho2014learning}, attention models \cite{vaswani2017attention} and BERT models \cite{devlin2018bert} thus proving the potential it holds in the machine learning area.

The work of~\newcite{nickel2017poincare} introduces and explores the potential of hyperbolic embedding by using an n-dimension Poincar\'e ball. The research work compares the hyperbolic and Euclidean embeddings for a complex latent data structure and comes to the conclusion that hyperbolic embedding surpasses the Euclidean embedding in effectivity. Inspired by the above results, both~\newcite{leimeister2018skip} and~\newcite{dhingra2018embedding} have extended the methodology introduced by~\newcite{nickel2017poincare}. \newcite{leimeister2018skip} have developed a hyperbolic word embedding using the skip-ngram negative sampling architecture taken from Word2vec. In lower embedding dimensions, the developed model performs better in comparison to its Euclidean counterpart. The work of~\newcite{dhingra2018embedding} uses re-parameterization to extend the Poincar\'e embedding, in order to learn the embedding of arbitrarily parameterized objects. The framework thus created is used to develop word and sentence embeddings. In our research, we will be following the footsteps of the above papers.

When considering the usage of hyperbolic embeddings in a practical context, the work of~\newcite{lu2019learning,lu2020exploiting} can be examined. The research by~\newcite{lu2019learning} improves the state-of-the-art model used to predict ICU (intensive care unit) re-admissions and surpasses the accepted benchmark used to predict in-hospital mortality using hyperbolic embedding of Electronic Health Records, while the work of~\newcite{lu2020exploiting} introduces a novel network embedding method which is capable of maintaining the consistency of the node representation across two views of networks, thus emphasizing the capabilities of hyperbolic embeddings. To the best of our knowledge, hyperbolic embeddings have not been previously applied to Sinhala content. Therefore, this research may reveal novel insight regarding hyperbolic embedding and its effectivity in sentiment analysis. 

In the research work of~\newcite{senevirathne2020sentiment}, capsule-B model \cite{zhao2018investigating} is crowned as the state-of-the-art model for the Sinhala sentiment analysis. In this work, a set of deep learning models are tested for the ability to predict the sentiment of Sinhala news comments. The GRU \cite{chung2014empirical} model with a CNN \cite{wang2016combination} layer which is used for the testing of each embedding in this work is taken from the aforementioned research. Furthermore, the work of~\newcite{weeraprameshwara2022sentiment} has extended the idea and tested the same set of deep learning models with the addition of sentiment analysis models introduced in the work of~\newcite{jayawickrama2021seeking} using the Facebook data set which is used in this research work. According to their results, the 3 layer stacked BiLSTM model \cite{zhou2019sentiment} outshines as the state-of-the-art model. 

\section{Methodology}
\label{sec:meth}

In order to test the feasibility of two-tiered word representation as a means of representing Sinhala text in the sentiment analysis domain, a series of experiments were conducted as described in the following subsections. 


\subsection{Data Set}
\label{sec:data}

The data set used for the project is extracted from the work of~\newcite{wijeratne2020sinhala}, which contains 1,820,930 Facebook posts from 533 Facebook pages popular in Sri Lanka over the time window of 2010 to 2020. The research work has produced two cleaned corpora and a set of stop words for the given context. The larger corpus among them consists of a total of 28 to 29 million words. The data set covers a wide range of subjects such as politics, media, and celebrities. Table~\ref{Table:Fields} illustrates the fields taken from the data set for the embedding development, model training and testing phases.


\begin{table}[!htb]
\renewcommand{\arraystretch}{1.1}
\begin{tabular}{ |c|c|c|c|c|} 
\hline
 \textbf{Field Name} & \textbf{Total Count} & \textbf{Percentage(\%)} \\ 
 \hline
 Likes & 312,282,979 & 93.58\\
 Loves & 10,637,722 & 3.19\\
 Wow & 1,633,255 & 0.49 \\
 Haha & 5,377,815 & 1.61\\
 Sad & 2,611,908 & 0.78\\
 Angry & 1,158,182 & 0.35\\
 Thankful & 12,933 & 0.00\\
 \hline
\end{tabular}
\caption{The counts and percentages of the reactions in the Facebook data set}
\label{Table:Fields}
\end{table}

\subsection{Preprocessing}


Even though there are two preprocessed corpora introduced through the work of~\newcite{wijeratne2020sinhala}, the raw data set was used for this research with the objective of preprocessing it to suit our requirements. As such, numerical content, URLs, email addresses, hashtags, words in other languages except for Sinhala and English, and excessive spaces were removed from the text. While the focus of this study is colloquial Sinhala, English is included in the data set as the two languages are often codemixed in colloquial use. Codemixing of Sinhala with other languages is much less in comparison. Furthermore, stop words were removed from the text as well, as recommended by~\newcite{wijeratne2020sinhala}. Posts with no textual content after thus preprocessing as well as posts with no reaction annotations were also removed as they yield no value in the annotation stage. The final preprocessed data set consists of Sinhala, English, and Sinhala-English code mixed content, adding up to a total of 542,871 Facebook posts consisting of 8,605,849 words.

\subsection{Annotation}
\label{sec:anno}
Since the procedure followed in the model development is supervised learning, the data set needed to be annotated \cite{schapire2012foundations}. It is quite a considerable challenge to obtain sufficiently large annotated data sets for resource-poor languages like Sinhala thus Facebook data set is ideal for the given scenario as the Facebook posts are pre-annotated by Facebook users using Facebook reactions. Though this is not an expert annotation, it can be considered as an effective means of community annotation as the collective opinion of a large number of Facebook users is represented by the reaction annotation \cite{pool2016distant,freeman2020measuring,graziani2019jointly,jayawickrama2021seeking}.   

\begin{figure}[!htb]
\centering
\includegraphics[width=0.35\textwidth]{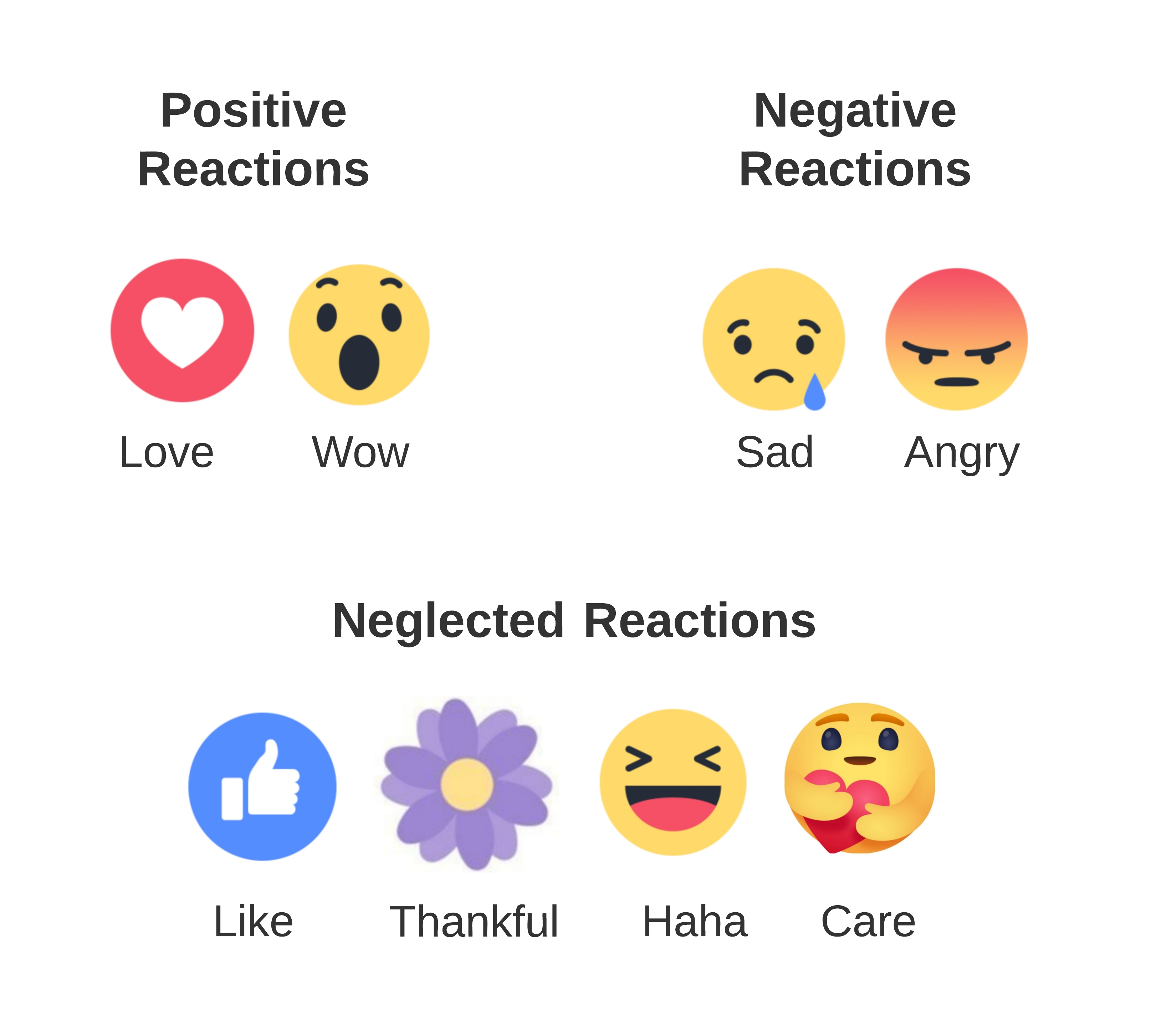}
    \caption{Reaction categorization for the annotation}
    \label{fig:reactions}
\end{figure}

A binary classification method which was introduced through the work of~\newcite{senevirathne2020sentiment} and further improved for Facebook data by~\newcite{weeraprameshwara2022sentiment} is used in this research as the annotation schema which is illustrated in the figure~\ref{fig:reactions}. Here, the Facebook reactions are divided into two classes; positive reactions and negative reactions. The reactions \textit{love} and \textit{wow} are considered as positive reactions while \textit{sad} and \textit{angry} are classified as negative reactions. The reactions \textit{like} and \textit{thankful} have been excluded as they are outliers in the data set with respect to the other reactions. The \textit{like} is the de facto reaction given by the users and it does not yield a valid sentiment. The \textit{thankful} reaction has appeared in a small time period making the presence insignificant compared to other reactions (only 0.00003\% of the total reaction count). The \textit{haha} reaction is also excluded due to the contradicting nature of its use cases \cite{jayawickrama2021seeking}. The \textit{care} reaction is not included in this data set as it was first introduced to the platform in 2020 \cite{lyles2022}, after the creation of the data set.

\subsection{Word Embeddings}
\label{sec:wordemb}
The final vector representation of Facebook posts consists of two major elements: word embeddings and sentence embeddings.

Word embeddings are used both as the first tier of the two-tiered embedding systems and as the basic one-tiered embedding systems used in the form of a benchmark against which the performance of two-tiered embedding systems would be compared. The performance of both Euclidean and hyperbolic word embeddings has been thus evaluated in this research.

\subsubsection{Euclidean Word Embeddings}
\label{sec:euemb}
For the purpose of representing words in the Euclidean space; fastText, Word2vec, and GloVe word embedding techniques were utilized. Word vectors consisting of 200 dimensions were created using each of the aforementioned models and a window size of 40 was picked based on the work of~\newcite{senevirathne2020sentiment,weeraprameshwara2022sentiment} which precedes this research. 

\subsubsection{Hyperbolic Embeddings}
\label{sec:hypemb}
The hyperbolic space exhibits different mathematical properties in comparison to the Euclidean space. Due to its inherent properties, the Euclidean space struggles to model a latent hierarchy. This issue could be addressed by mapping the embedding into a higher dimension \cite{nickel2017poincare}. However, this may lead to sparse data mapping, causing the curse of dimensionality to affect the performance. This may induce adverse effects such as causing the machine learning model to overfit by the data and using a high memory capacity for computations and storage.

The hyperbolic space has caught the attention of researchers as a plausible solution to such issues encountered in using the Euclidean space for modeling complex structures. The unique feature of this mathematical model is that the space covered by an n-ball in an n-dimensional hyperbolic space increases exponentially with the radius. In contrast to the Euclidean space where the space covered by an n-ball remains restricted by the $n^{th}$ power of the radius, the hyperbolic space could easily handle complex models such as tree-like structures within a limited dimensionality.

The distance ($D$) between two vectors ($i$ and $j$) in the hyperbolic space can be calculated as shown in equation~\ref{Eq:hdis}.

\begin{equation}
     D_{(i,j)} = \arccosh{(1+\frac{2\lvert\lvert i-j\rvert\rvert^2}{(1-\lvert\lvert i\rvert\rvert^2)(1-\lvert\lvert j\rvert\rvert^2)})}
\label{Eq:hdis}
\end{equation}




Since both the circumference and the area of a hyperbolic circle grow exponentially with the radius, the hyperbolic space has the capability to effectively store a complex latent hierarchy of data using a much lower number of dimensions than the Euclidean space would require to store the exact same structure. 

\begin{figure}[!htb]
\centering
\includegraphics[width=0.35\textwidth]{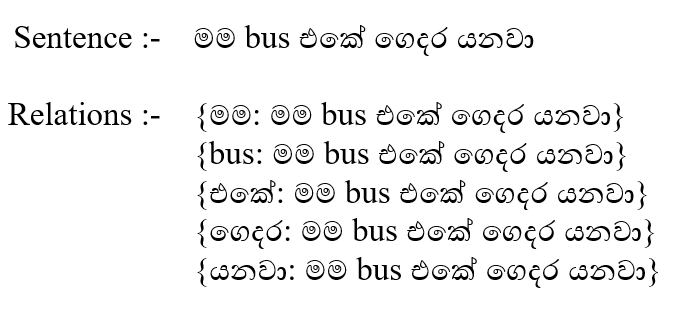}
    \caption{Examples of parsing mechanism used for the hyperbolic embeddings where each word is matched to the sentence}
    \label{fig:sent}
\end{figure}

In order to create hyperbolic word embeddings, the data set should be reformed in such a manner that the syntactic structure of data is highlighted. However, an adequate language parser for Sinhala does not currently exist \cite{de2019survey}. Using parsers dedicated to the English language is also unfitting since the underlying grammatical structure of Sinhala is significantly different from that of English. Furthermore, for codemixed colloquial data present in this data set, grammatical structures of both Sinhala and English languages would have to be taken into consideration. Therefore, the parsing mechanism shown in figure\ref{fig:sent} is used to generate word tokens. A total of 8605849 tokens have been thus generated.

\begin{figure}[!htb]
\centering
\includegraphics[width=0.32\textwidth]{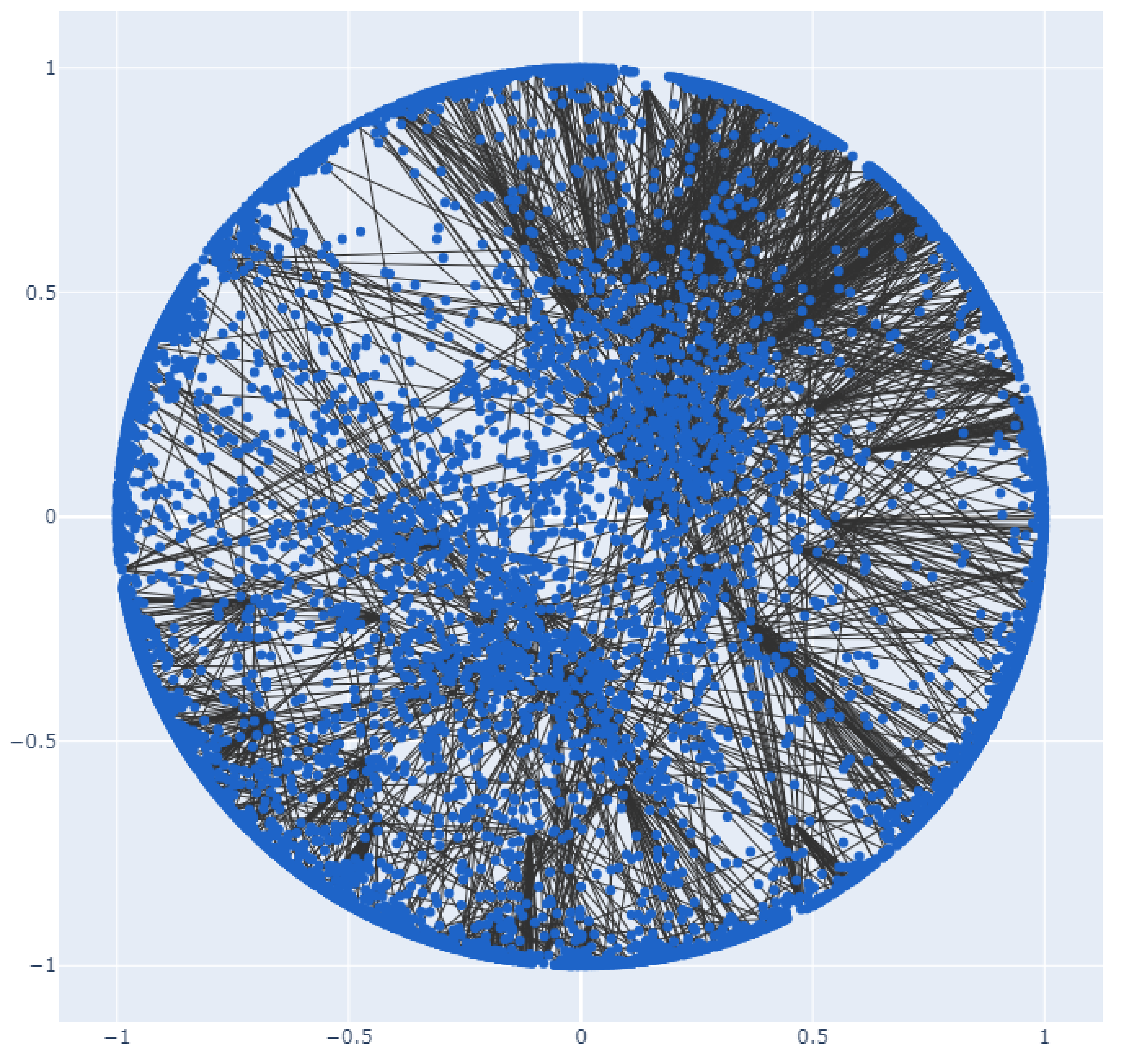}
    \caption{Poincar\'e word embedding done on Facebook data set}
    \label{fig:poincare}
\end{figure}

The two-dimensional illustration of the Poincar\'e ball after training with the Facebook data set is shown in the figure~\ref{fig:poincare}. Each node represents a word in the figure and each edge represents the connection between words. Here for the illustration purposes, only a thousand nodes are shown and the dimension is projected from 200 to 2.

\begin{figure}[!htb]
\centering
\includegraphics[width=0.495\textwidth]{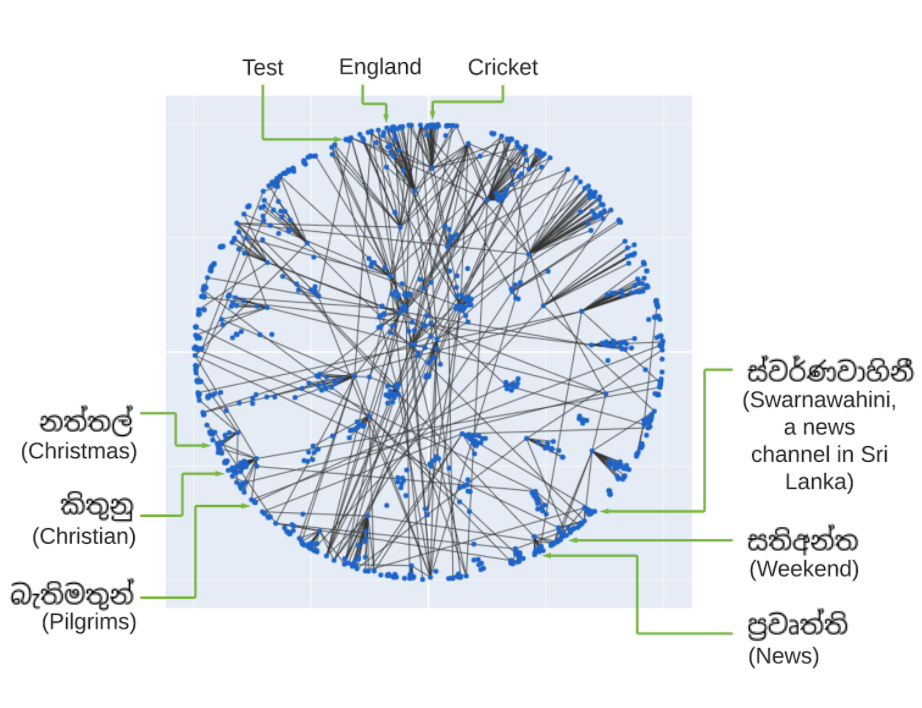}
    \caption{Word clustering in the Poincar\'e embedding. The meaning of the Sinhala words are given in the brackets}
    \label{fig:poinwords}
\end{figure}

The clustering of semantically related words in the Poincar\'e embedding is shown in the figure~\ref{fig:poinwords}. A set of words related to cricket sport is clustered in the top left corner while a set of Sinhala words related to Christianity is clustered in the bottom left. A cluster which represents news-related terms is formed in the bottom right corner. With this evidence, we can safely assume that the hyperbolic space has the capability to store a complex latent hierarchy such as the semantic relation of words. 

\subsection{Sentence Embeddings}
\label{sec:sentemb}

Sentence embeddings are used as the second tier of the two-tiered embedding models. Basic pooling methods as well as the sequence to sequence model are used to generate sentence embeddings by using the word embedding of each word in a sentence. For both Euclidean space and hyperbolic space embeddings, the sentence embeddings are generated in a similar fashion as described below.



\subsubsection{Pooling}
\label{sec:pool}
Sentence embeddings have been created with three different pooling mechanisms for each of fastText, Word2vec, GloVe, and hyperbolic word embeddings; namely, max pooling, min pooling, and avg pooling. Pooling embeddings will be considered as baseline sentence embeddings against which the performance of the sequence to sequence model is compared.







\subsubsection{Sequence to Sequence Model}
\label{sec:seq2seq}

This sentence embedding mechanism follows the sequence to sequence model introduced through the work of~\newcite{sutskever2014sequence}, referred to as the seq2seq model from here onwards. 



The data set is randomly shuffled and a subset consisting of 400,000 data rows is used for training the encoder, decoder units. 

In the original model, the encoder accepts a set of vectors which consists of the word embedding of each word in a sentence followed by the $<EOS>$ token as input and returns a context vector as the output. In order to train the model, the decoder is fed with the context vector from the encoder, with the objective of getting the $<SOS>$ token followed by the translated sentence as the final output. For our research, the output expected from the decoder is the same sentence that has been inputted into the encoder. For a given sentence, the word embedding of each word in the sentence is inputted into the Recurrent Neural Network encoder, which has a hidden layer similar in dimensions to the word embedding. Since the expected output from the RNN decoder is also the same sentence, the context vector (output of the encoder) can be considered as the sentence embedding that we are seeking. 

Different sentence embeddings are thus generated using both Euclidean and hyperbolic word embeddings as inputs to the seq2seq model. For each type of word embeddings, the RNNs inside the encoder and decoder are also modified to generate different sentence embeddings. Here, GRU \cite{chung2014empirical}, LSTM \cite{hochreiter1997long}, and simple RNN models are used. The architecture of the GRU seq2seq model has been inspired by the model introduced through the work of~\newcite{cho2014learning}.

Furthermore, two different decoder structures have been used to train the seq2seq model. A simple decoder which functions as explained above and a decoder with the attention mechanism introduced in the work of~\newcite{vaswani2017attention} are thus utilized. Both models use a teacher forcing value of 0.5 with the objective of performing better at the prediction task \cite{lamb2016professor}. 

The squared L2 norm between the predicted word embedding and the actual word embedding is used as the loss function for Euclidean embeddings. Equations~\ref{Eq:predict} shows embedding value of the $i$th sentence which is calculated by summing up all the word embeddings in the predicted sequence of word embeddings. The symbol $n$ is the length of the longest sentence which may vary for the selected data set. $WE_k$ is the value of each dimension in the 200 dimension word embedding. Equation ~\ref{Eq:actual} calculates the value of $i$th true word embedding sequence ($TV_i$) which is the summation of the word embeddings of True word sequence. Then in the equation~\ref{Eq:eloss}, the squared L2 norm ($Err$) is calculated. $n$ denotes the number of data items used. The procedure follows for both Euclidean and hyperbolic space embeddings.

\begin{equation}
    PV_i={\sum_{j=1}^{n}\sum_{k=1}^{200}WE_K}\\
\label{Eq:predict}
\end{equation}


\begin{equation}
    TV_i={\sum_{j=1}^{n}\sum_{k=1}^{200}WE_K}\\
\label{Eq:actual}
\end{equation}
\begin{equation}
    Err=(1/n){\sum_{i=1}^{n}(PV_i - TV_i)^2}
\label{Eq:eloss}
\end{equation}



\subsection{Testing}
\label{sec:test}
To the extent of our knowledge, there does not exist a well known or effective benchmark to test the performance of Sinhala sentence embeddings. Therefore, the GRU RNN model with a CNN layer introduced by the work of~\newcite{chung2014empirical,senevirathne2020sentiment} is used to test each embedding. The function of this model is to understand the sentimental reactions of Facebook users to Facebook posts and thus classify each post as either positive or negative based on its prediction of the sentimental reaction of users to that post. The classification of the Facebook posts was done as explained in the section~\ref{sec:anno}. 

As mentioned above since the scarcity of a large enough data set for Sinhala language to train deep learning models, the same Facebook data set is used for the model training purpose. However, a different set of Facebook posts are used in order to avoid repetition of the data set and a total of 200,000 posts were used for the training purpose. The holdout method was used with data set splits into the 8:1:1 ratio for train, validate, and test sets. Tests were run multiple times and the average performance measures were recorded.



\section{Results}
\label{sec:res}
The results obtained by training the models only using word embeddings are displayed in table~\ref{Table:Words}. Here, the row fastText(Sinhala News Comments) taken from the work of~\newcite{weeraprameshwara2022sentiment} is used as a benchmark against which the performance measures of the other word embeddings are compared. There, the Facebook data set was embedded using the fastText word embeddings trained with the Sinhala News Comments data set introduced through the work of~\newcite{senevirathne2020sentiment}, while the latter rows display the results of embedding the Facebook data set with word embeddings trained with the Facebook data set itself.

As the table portrays, using the Facebook data set containing 542,871 preprocessed Facebook posts, which is much larger in size than the Sinhala News Comments data set with 15,000 Sinhala News comments, to develop the word embeddings has resulted in a comparatively higher F1 score.

\begin{table*}[!htb]
\centering
\begin{tabular}{|p{7.5cm}|c|c|c|c|}
\hline
 \multirow{2}{*}{\textbf{Word Embedding}} &  \multicolumn{4}{c|}{\textbf{Performance Measures}}\\
 \hhline{~----}
 & \textbf{Accuracy} & \textbf{Precision} &\textbf{Recall} & \textbf{F1} \\
 \hline
 fastText (Sinhala News comments) \cite{weeraprameshwara2022sentiment} & 81.17 & 81.17 & 81.57 & 81.37 \\
  \hline
 Word2vec & 83.47 & 83.65 & 83.47 & 83.56 \\
 GloVe & 82.09 & 81.91 & 82.65 & 82.28\\
 fastText & \textbf{83.76} & \textbf{83.76} & \textbf{83.76} & \textbf{83.76} \\
 Hyperbolic & 82.78 & 82.11 & 83.58 & 82.84\\
 \hline
\end{tabular}
\caption{Word Embedding results}
\label{Table:Words}
\end{table*}

The results of each embedding in the two-tiered structure are shown in table~\ref{Table:results}. The first column presents the word embedding method used while the second column depicts the sentence embedding method utilized and the rest of the columns are used to present the performance measures. The best performance measures from each word embedding category are highlighted.

\FPeval{\MAXW}{86.72}
\FPeval{\MAXG}{86.22}
\FPeval{\MAXF}{87.49}
\FPeval{\MAXH}{85.77}

\FPeval{\MINW}{86.64}
\FPeval{\MING}{85.99}
\FPeval{\MINF}{87.52}
\FPeval{\MINH}{85.11}

\FPeval{\AVGW}{87.01}
\FPeval{\AVGG}{85.93}
\FPeval{\AVGF}{87.93}
\FPeval{\AVGH}{85.47}

\FPeval{\GSW}{85.75}
\FPeval{\GSG}{85.16}
\FPeval{\GSF}{86.23}
\FPeval{\GSH}{86.13}

\FPeval{\GAW}{87.29}
\FPeval{\GAG}{85.12}
\FPeval{\GAF}{88.04}
\FPeval{\GAH}{86.54}

\FPeval{\LSW}{86.01}
\FPeval{\LSG}{85.16}
\FPeval{\LSF}{86.60}
\FPeval{\LSH}{85.81}

\FPeval{\LAW}{86.53}
\FPeval{\LAG}{85.12}
\FPeval{\LAF}{87.72}
\FPeval{\LAH}{86.30}

\begin{table*}[!htb]
\centering

\begin{tabularx}{\textwidth}{|l|l||*{4}{Y|}}
\hline
\multicolumn{2}{|c||}{\textbf{Embedding level}} &
 \multicolumn{4}{c|}{\textbf{Performance Measures}} \\
 \hline
\textbf{Word} & \textbf{Sentence} & \textbf{Accuracy} & \textbf{Precision} & \textbf{Recall} & \textbf{F1 Score}\\
 \hline
 \multirow{7}{*}{Word2vec} 
 & Max Pooling & 77.23 & 80.06 & 94.59 & \MAXW \\
 & Min Pooling & 77.29 & \textbf{81.55} & 92.41 & \MINW\\
 & Avg Pooling & 77.44 & 81.43 & 93.41 & \AVGW\\
 & Seq2seq GRU & 75.86 & 76.74 & 97.14 & \GSW\\
 & Seq2seq GRU with attention & \textbf{79.12} & 79.72 & 96.45 & \textbf{\GAW}\\
 & Seq2seq LSTM & 75.97 & 76.17 & \textbf{98.76} & \LSW\\
 & Seq2seq LSTM with attention & 77.42 & 77.86 & 97.36 & \LAW\\
 \hline
 
 \multirow{7}{*}{GloVe} 
 & Max Pooling & 75.63 & 77.74 & 96.79 & \textbf{\MAXG}\\
 & Min Pooling & 75.34 & \textbf{78.68} & 94.81 & \MING\\
 & Avg Pooling & \textbf{76.11} & 76.90 & 97.38 & \AVGG\\
 & Seq2seq GRU & 74.23 & 74.15 & \textbf{100.00} & \GSG\\
 & Seq2seq GRU with attention & 74.23 & 74.09 & \textbf{100.00} &  \GAG\\
 & Seq2seq LSTM & 74.23 & 74.15 & \textbf{100.00} & \LSG\\
 & Seq2seq LSTM with attention & 74.23 & 74.09 & \textbf{100.00} & \LAG\\
 \hline
 
 \multirow{7}{*}{fastText} 
 & Max Pooling & 79.93 & 81.23 & 94.78 & \MAXF\\
 & Min Pooling & 79.80 & 82.49 & 93.22 & \MINF\\
 & Avg Pooling & \textbf{80.86} & \textbf{82.55} & 94.07 & \AVGF\\
 & Seq2seq GRU & 78.12 & 80.90 & 92.33 & \GSF\\
 & Seq2seq GRU with attention & 80.61 & 81.31 & 96.00 & \textbf{\GAF}\\
 & Seq2seq LSTM & 79.00 & 82.12 & 91.59 & \LSF\\
 & Seq2seq LSTM with attention & 80.31 & 80.06 & \textbf{96.98} & \LAF\\
 \hline
 
 \multirow{7}{*}{Hyperbolic} 
 & Max Pooling & 76.71 & 77.54 & 95.95 & \MAXH\\
 & Min Pooling & 76.11 & 77.68 & 94.11 & \MINH\\
 & Avg Pooling & 77.00 & 77.31 & 95.56 & \AVGH\\
 & Seq2seq GRU & 76.38 & 77.09 & \textbf{97.57} & \GSH\\
 & Seq2seq GRU with attention & \textbf{77.31} & \textbf{78.31} & 96.70 & \textbf{\GAH}\\
 & Seq2seq LSTM & 76.48 & 77.91 & 95.49 & \LSH\\
 & Seq2seq LSTM with attention & 77.19 & 78.22 & 96.24 & \LAH\\
 \hline
\end{tabularx} 
\caption{Performance measures of each embedding}
\label{Table:results}
\end{table*}

The best F1 score is produced by the two-tiered embedding which uses fastText as the word embedding and the seq2seq model with GRU RNNs and attention layer as the sentence embedding while the second-best F1 is scored by the fastText embedding with average pooling. For each of the sentence embedding methods, the highest F1 score is produced by pairing with fastText word embeddings. fastText embeddings have resulted in a better F1 score in the one-tiered embeddings as well. Thus, we can conclude that fastText is the word embedding schema which provides the best performance in this context.

Upon taking the word embedding categories into consideration, Word2vec embeddings provide the second-best results, with performance scores slightly lower than those of fastText. The ranking of F1 scores achieved by hyperbolic and GloVe embeddings seem to be highly dependent on the type of sentence embedding used. However, the best F1 score obtained by hyperbolic embeddings, which was by pairing with the seq2seq model with GRU encoder and decoder units including an attention layer, is higher than the best F1 score GloVe embeddings have achieved upon pairing with max pooling sentence embeddings. It should be noted that the structure of data utilized here may not be optimal for hyperbolic embeddings.

In sentence embeddings, the performance of seq2seq model with GRU encoder, decoder units and an attention layer tend to surpass other sentence embedding models except when the word embedding utilized is GloVe. Nonetheless, stripping off the attention layer brings the performance of the seq2seq model with LSTM encoders and decoders to a level higher than that obtained with GRU encoder and decoder units, with the exception of the case where hyperbolic word embeddings are utilized.

Furthermore, there is a clear improvement in the performance scores of the seq2seq model when the attention layer is applied to the decoder. However, when the attention layer is not applied, pooling embeddings manage to perform better than seq2seq models except when hyperbolic word embeddings are utilized. The reason for this exception could be that the Euclidean pooling mechanisms used may not be the best fit for hyperbolic embeddings.

\section{Conclusion}
\label{sec:conc}

Comparing tables~\ref{Table:Words} and~\ref{Table:results} makes it evident that there is a clear improvement in performance when two-tiered embedding systems are used, in contrast to simply using a single tier of word embeddings. The possibility of sentence embeddings used in two-tiered embedding systems to enable the models to consider the syntax of sentences could be the reason for this improvement. When word embeddings of Sinhala Facebook posts are directly fed to a sentiment analysis model, the model is likely to see the Facebook posts as merely an unorganized set of words instead of an organized set of sentences.

In addition, the results displayed in table~\ref{Table:results} exhibit the use of the two-tiered embedding system that combines fastText word embeddings and seq2seq sentence embeddings with GRU encoder and decoder units as well as an attention layer has given rise to the best performance measures. Although Word2vec embeddings follow closely behind in performance, they have failed to surpass fastText, possibly due to the inability of the embedding system to consider the internal structure of words, which the fastText embedding system by nature is capable of \cite{bojanowski2017enriching,joulin2016bag}.

Though the hyperbolic space has an advantage over the Euclidean space due to its ability to effectively represent complex hierarchical data structures \cite{nickel2017poincare}, fastText and Word2vec embeddings have outperformed hyperbolic embeddings in this research. The reason for this could be the lack of potent parsing tools for the Sinhala language \cite{de2019survey}. To obtain the optimum performance from hyperbolic embeddings, an effective hierarchical structure such as sentence structures identified via parsing is required. The simple $[word, sentence]$ relation structure used in this research may not be sufficient for this. Furthermore, the pooling techniques also fail to be on par with the seq2seq model, possibly due to the fact that the vectors generated by applying Euclidean pooling mechanisms on hyperbolic embeddings do not always fall within the space of the Poincar\'e ball.

Another noteworthy fact is that the GloVe embeddings tend to underperform in comparison to the other word embeddings models used in this research. Unlike resource-rich languages such as English, no pre-trained GloVe models exist for the Sinhala language. This could hinder the ability of GloVe embeddings to achieve their full potential.

Thus, it can be concluded that though a robust embedding model for Sinhala that is applicable across all domains may not be currently available, it could be possible to develop an effective embedding system that would at least be potent within the domain of the training data set by applying a two-tiered embedding model such as the seq2seq sentence embeddings with GRU encoders and decoders stacked on top of fastText word embeddings on a sufficiently large data set.

\section{Future Work}
\label{sec:future}

This research is related to the work of~\newcite{jayawickrama2021seeking} and as the final goal, a Facebook reaction prediction tool for colloquial Sinhala text will be developed and the word representations developed in this project will be used for that tool.

The data set contains both Sinhala and English text since our aim is to develop a word representation for colloquial Sinhala text which consists of English and Sinhala code-mixed content. However, a pure Sinhala embedding can be generated in the future. 

Furthermore, Poincar\'e embeddings could be developed for Sinhala text with the use of a proper parser to identify sentence structures though developing a reasonable parser for colloquial text will be a challenge.

Though this research only considers sentiment analysis for the Sinhala language, the applicability of the two-tiered embedding systems discussed in other areas of natural language processing as well as for other resource-poor languages could be tested as well.


\bibliography{bibliography}
\bibliographystyle{acl_natbib}

\end{document}